\documentclass[10pt,twocolumn,letterpaper]{article}

\usepackage{cvpr}              

\usepackage{graphicx}
\usepackage{amsmath}
\usepackage{amssymb}
\usepackage{booktabs}
\usepackage{MnSymbol}
\usepackage{algorithm}
\usepackage[noend]{algpseudocode}
\usepackage{colortbl}
\usepackage{multirow}

\newcommand{\algname}{fairness-aware adversarial perturbation}
\newcommand{\algabbr}{FAAP}

\definecolor{mygray}{gray}{.9}

\usepackage[pagebackref,breaklinks,colorlinks]{hyperref}
\usepackage{lipsum}

\usepackage[capitalize]{cleveref}
\crefname{section}{Sec.}{Secs.}
\Crefname{section}{Section}{Sections}
\Crefname{table}{Table}{Tables}
\crefname{table}{Tab.}{Tabs.}

\def\cvprPaperID{6962} 
\def\confName{CVPR}
\def\confYear{2022}

\begin{document}

\title{Fairness-aware Adversarial Perturbation Towards Bias Mitigation for \\Deployed Deep Models}

\author{Zhibo Wang$^{\dagger,\ddagger}$, Xiaowei Dong$^{\dagger}$, Henry Xue$^{\star}$, Zhifei Zhang$^{\sharp}$, Weifeng Chiu$^{\star}$, Tao Wei$^{\star}$, Kui Ren$^{\ddagger}$\\
$^{\dagger}$School of Cyber Science and Engineering, Wuhan University, P. R. China\\
$^{\ddagger}$School of Cyber Science and Technology, Zhejiang University, P. R. China\\
$^{\star}$Ant Group, \quad
$^{\sharp}$Adobe Research\\
{\tt\small \{zhibowang, kuiren\}@zju.edu.cn, xwdong@whu.edu.cn, \{weifeng.qwf, lenx.wei\}@antgroup.com,}\\
{\tt\small  zzhang@adobe.com, gkn1fexxx@gmail.com}
}
\maketitle
\newcommand\blfootnote[1]{%
\begingroup
\renewcommand\thefootnote{}\footnote{#1}%
\addtocounter{footnote}{-1}%
\endgroup
}
\blfootnote{This manuscript was accepted by CVPR 2022.}

\begin{abstract}
    Prioritizing fairness is of central importance in artificial intelligence (AI) systems, especially for those societal applications, \eg, hiring systems should recommend applicants equally from different demographic groups, and risk assessment systems must eliminate racism in criminal justice. Existing efforts towards the ethical development of AI systems have leveraged data science to mitigate biases in the training set or introduced fairness principles into the training process. For a deployed AI system, however, it may not allow for retraining or tuning in practice. By contrast, we propose a more flexible approach, \ie, fairness-aware adversarial perturbation (\algabbr), which learns to perturb input data to blind deployed models on fairness-related features, \eg, gender and ethnicity. The key advantage is that \algabbr{} does not modify deployed models in terms of parameters and structures. To achieve this, we design a discriminator to distinguish fairness-related attributes based on latent representations from deployed models. Meanwhile, a perturbation generator is trained against the discriminator, such that no fairness-related features could be extracted from perturbed inputs. Exhaustive experimental evaluation demonstrates the effectiveness and superior performance of the proposed \algabbr{}. In addition, \algabbr{} is validated on real-world commercial deployments (inaccessible to model parameters), which shows the transferability of \algabbr, foreseeing the potential of black-box adaptation.   
\end{abstract}

\section{Introduction}

AI systems have been widely deployed in many high-stakes applications, \eg, face recognition~\cite{b1,b2}, hiring process~\cite{b3,b4}, health care~\cite{b5}, \etc. However, some existing AI systems are found to treat individuals unequally based on protected attributes, \eg, ethnicity, gender, and nationality. Such biases are referred to as unfairness. For instance, Amazon realized that their automatic recruitment system presents skewness between male and female candidates~\cite{b6}, \ie, male candidates are with higher probability to be hired as compared to female candidates. The COMPAS, which is an assessment system of recidivating risk, is found to have racial prejudice~\cite{b8}. Such unfairness has been a subtle and ubiquitous nature of AI systems, thus it is non-trivial to mitigate the unfairness, ideally without touching the deployed models.

Many works have been proposed to mitigate unfairness/biases, which can be divided into three categories according to the stage de-biasing is applied, \ie, pre-processing, in-processing, and post-processing. From the perspective of pre-processing, \cite{b9,b10,b11,b12,b20} mitigated biases in the training dataset, thus mitigating the bias during training the model. For the in-processing methods, \cite{b14,b28,b56} introduced fairness-related penalties into the learning process to train a fairer model. These methods need to retrain or fine-tune the target models, while these are unsuitable if the models are deployed without access to their training set. \cite{b18} proposed a boosting method to post-process a deployed deep learning model to produce a new classifier that has equal accuracy in different people groups. However, \cite{b18} needs to replace the final classifier and cannot ensure statistical and predictive parity, \eg, individuals in different groups are equally treated in prediction.

To the best of our knowledge, existing works are not suitable to improve fairness at the inference phase without changing the deep model. Therefore, it is imperative to propose a practical approach to mitigate the unfairness of deployed models without changing their parameters and structures. Since deep models tend to learn spurious correlations between protected attributes and target labels from training data, \eg, the race may correlate to criminal risk, the key to mitigating unfairness is to break such correlation. As we assume not modifying the model, the main challenge of achieving this goal is how to prevent the deployed model from extracting fairness-related information from inputs. Intuitively, the only thing we could modify is the input data during the inference stage of deployed models, \ie, perturbing the inputs such that the model cannot recognize those protected attributes.

\begin{figure}[!t]
\centering
\includegraphics[width=1\linewidth]{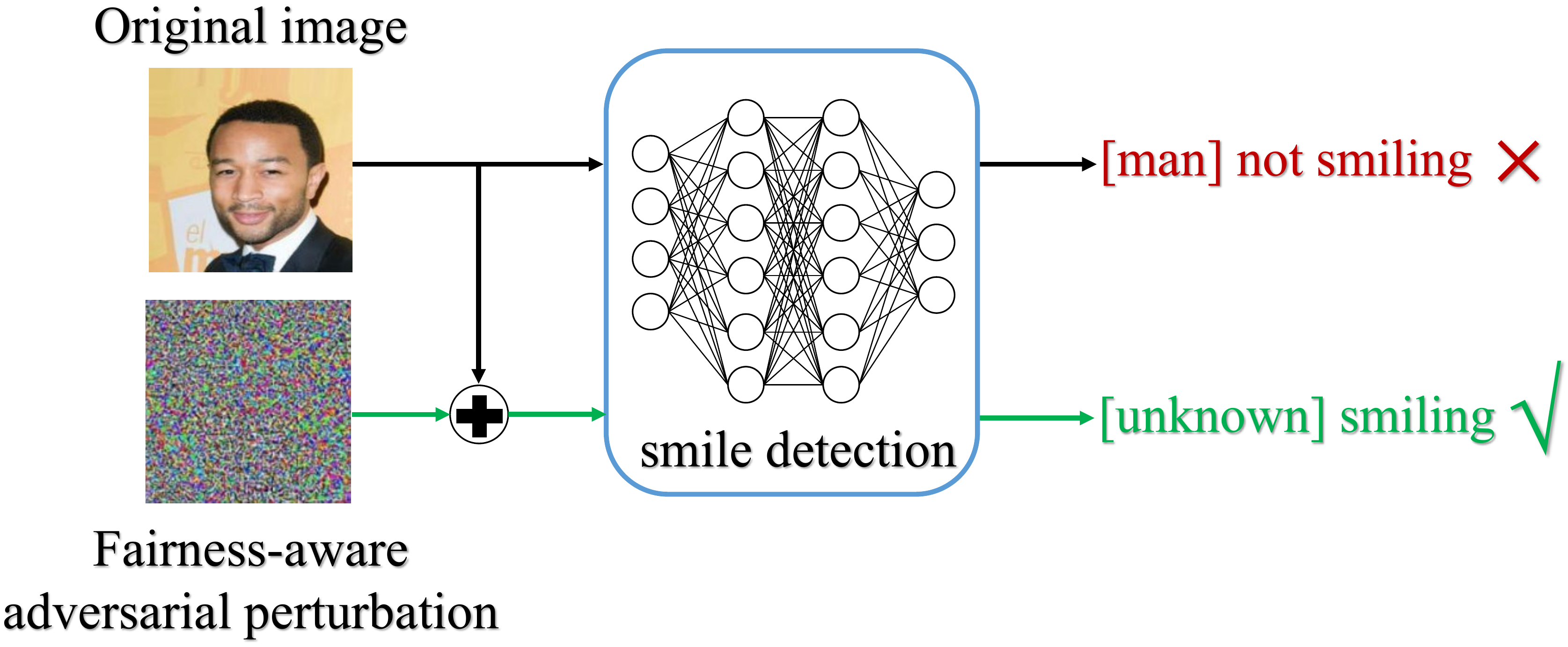}
\vspace{-2mm}
\caption{The illustration on a smile detection model. Original image is falsely recognized due to model unfairness, \ie, tending to predict males as ``not smiling''. The fairness-aware adversarial perturbation generated by \algabbr{} helps the input image to hide the protected attribute and get fair treatment.}
\label{figure:fig1}
\vspace{-5mm}
\end{figure}

Based on the above idea, we propose the Fairness-Aware Adversarial Perturbation (FAAP), which learns to perturb input samples to blind deployed models on fairness-related features. As shown in Fig.~\ref{figure:fig1}, the deployed model can not distinguish the fairness-related feature (\eg, gender) from the perturbed input image. Therefore, the predictions will not correlate to the protected attributes. The key idea is that perturbations can remap samples to tightly distribute along the decision hyperplane of protected attributes in the model latent space, making them difficult to be distinguished. To achieve this, we train a generator to produce adversarial perturbation. During the training process, a discriminator is trained to distinguish the protected attributes from the representations of the model, while the generator learns to deceive the discriminator, thus generating fairness-aware perturbation that can hide the information of protected attributes from the feature extraction process. Extensive experimental evaluation demonstrates the superior performance of the proposed FAAP and shows the potential in the black-box scenario, \ie, mitigating unfairness of models without access to their parameters.

In summary, the main contributions of this paper are in three-folds:
\begin{itemize}
\item We give the first attempt to mitigate the unfairness from deployed deep models without changing their parameters and structures. This pushes the fairness research towards a more practical scenario. 
\item We propose the fairness-aware adversarial perturbation (FAAP), which designs a discriminator to distinguish fairness-related attributes based on latent representations from deployed models. Meanwhile, a generator is trained adversarially to perturb input data to prevent the deployed models from extracting fairness-related features. This design effectively decorrelates fairness-related/protected attributes from predictions.
\item Extensive experiments demonstrate the superior performance of the proposed \algabbr{}. In addition, evaluation on real-world commercial APIs shows the transferability of \algabbr, which indicates the potential of further exploring our method in the black-box scenario.
\end{itemize}

\section{Related work}
\label{section:Related work}

This section overviews related works on unfairness mitigation that could be roughly divided according to targeting stages, \ie, pre-processing (data pre-processing before training), in-processing (penalty design during training), and post-processing (prediction adjustment after training).

\textbf{Pre-processing} methods~\cite{b9,b10,b11} aim to mitigate biases in the training dataset, \ie, fairer training sets would train fairer models. Many methods have been proposed to de-bias training sets by fair data representation transformation or data distribution augmentation. Quadrianto~\etal~\cite{b9} used data-to-data translation to find middle-ground representation for different gender groups in training data, thus the model will not learn the tendency of gender. Ramaswamy~\etal~\cite{b10} generated paired training data to balance protected attributes, which would remove spurious correlation between target label and protected attributes. Zhang \etal~\cite{b11} proposed to generate adversarial examples to supplement the training dataset, balancing the data distribution over different protected attributes.

\textbf{In-processing} approaches~\cite{b27,b28,b56,b13,b14} introduce fairness principles into the training process, \ie, training models by specifially designed fairness penalties/constraints or adverasial mechinsm. 
Zafar~\etal~\cite{b27} proposed to maximize accuracy under disparate impact constraints to improve fairness in machine learning. Brian~\etal~\cite{b28} and Zhang~\etal~\cite{b56} enforced the model to produce fair outputs with adversarial training techniques by maximizing accuracy while minimizing the ability of a discriminator to predict the protected attribute. Yuji Roh~\etal~\cite{b13} provided a mutual information-based interpretation of an existing adversarial training-based method for improving the disparate impact and equalized odds. Sarhan~\etal~\cite{b14} imposed orthogonality and disentanglement constraints on the representation and forced the representation to be agnostic to protected information by entropy maximization, then the following classifier can make fair predictions based on learned representation. This line of research aims at getting a fairer model by explicitly changing the training procedure. Different from this line of work, our method is applied after the training process and can improve fairness without changing the deployed model.

\textbf{Post-processing} works~\cite{b17,b18} tend to adjust model predictions according to certain fairness criteria. Lohia~\etal~\cite{b17} proposed a post-processing algorithm that helps a model meet both individual and group fairness criteria on tabular data by detecting biases from model outputs and correspondingly editing protected attributes to adjust model predictions. However, this method needs to change protected attributes at the test time which is hard for computer vision applications. Michael~\etal~\cite{b18} proposed a method that can post-process a pre-trained deep learning model to create a new classifier, which has equal accuracy for people with different protected attributes. However, \cite{b18} needs to replace the final classifier, and equal sub-group accuracy can not ensure people in different groups have equal chance to get favorable predictions, \eg, unequal false positive rate and false negative rate. We borrow ideas from this line of research, but we improve fairness from the data side, instead of manipulating the model or its prediction.

\section{Preliminaries} \label{section:Preliminaries}

\subsection{Model fairness}
\label{subsection:Fairness definitions}

In this paper, we focus on visual classification models because of exhaustive academic efforts on them, as well as their broad industrial applications. Moreover, it is important to achieve equal treatment for people with different protected attributes, \eg, nationality, gender, and ethnicity. Therefore, demographic parity~\cite{b30} and equalized odds~\cite{b49} are adopted to measure model fairness.

In a binary classification task, \eg, criminal prediction, suppose target label $y \in \mathcal{Y} =\{-1,1\}$, protected attribute $z \in \mathcal{Z} = \{-1,1\}$, where $y=1$ is in favourable class (e.g., lower criminal tendency) and $z=1$ is in privileged group (e.g., Caucasian). 

\noindent \textbf{Definition 1} (Demographic Parity). If the value of $z$ does not influence assigning a sample to the positive class, i.e. model prediction $\hat{y}=1 \upmodels z$, then the classifier satisfies demographic parity:
\begin{equation} \label{equation:1}
    P(\hat{y}=1|z=-1) = P(\hat{y}=1|z=1)
\end{equation}

If a model satisfies demographic parity, samples in both the privileged and unprivileged groups have the same probability to be predicted as positive.

\noindent \textbf{Definition 2} (Equalized Odds). If the value of $z$ can not influence the positive outcome for samples given $y$, i.e. $\hat{y}=1 \upmodels z|y$,  then the classifier satisfies equalized odds:
\begin{equation} \label{equation:2}
P(\hat{y}=1|y,z=-1)=P(\hat{y}=1|y,z=1), y=\{-1,1\}
\end{equation}

Equalized odds means that positive output is statistically independent to the protected attribute given the target label. Samples in both the privileged and unprivileged groups have the same false positive rate and false negative rate.

\subsection{Adversarial examples}
Recent studies show that deep learning models are vulnerable to adversarial examples~\cite{b50}. Given a classification model $C(x)$, the goal of adversarial attacks is to find a small perturbation to generate an adversarial example $x^{\prime}$, to mislead classifier $C$. More specifically, there are two kinds of adversarial example attacks. For an input $x$ with ground label $y$, targeted attack will let $C(x^{\prime}) = y^{\prime}$ where $y^{\prime} \neq y$ is a label specified by the attacker. On the contrary, in an untargeted attack, an attacker will mislead the classification model as $C(x^{\prime}) \neq y$. Typically, the $l_p$ norm of the perturbation should be less than $\epsilon$, i.e. $\left\|x - x^{\prime}\right\|_p \leq \epsilon$. Many methods have been proposed to generate adversarial examples, such as PGD~\cite{b32}, CW~\cite{b34} and GANs based method~\cite{b35}.

\section{Fairness-aware adversarial perturbation} \label{section:Our method}
In this paper, we propose \algname{} (\algabbr) to mitigate unfairness born with deep models. This section will overview the proposed \algabbr{} and detail the design of network and loss functions. Finally, we will further discuss the training strategy of \algabbr.

\begin{figure}[!t]
\centering
\includegraphics[width=1.\columnwidth]{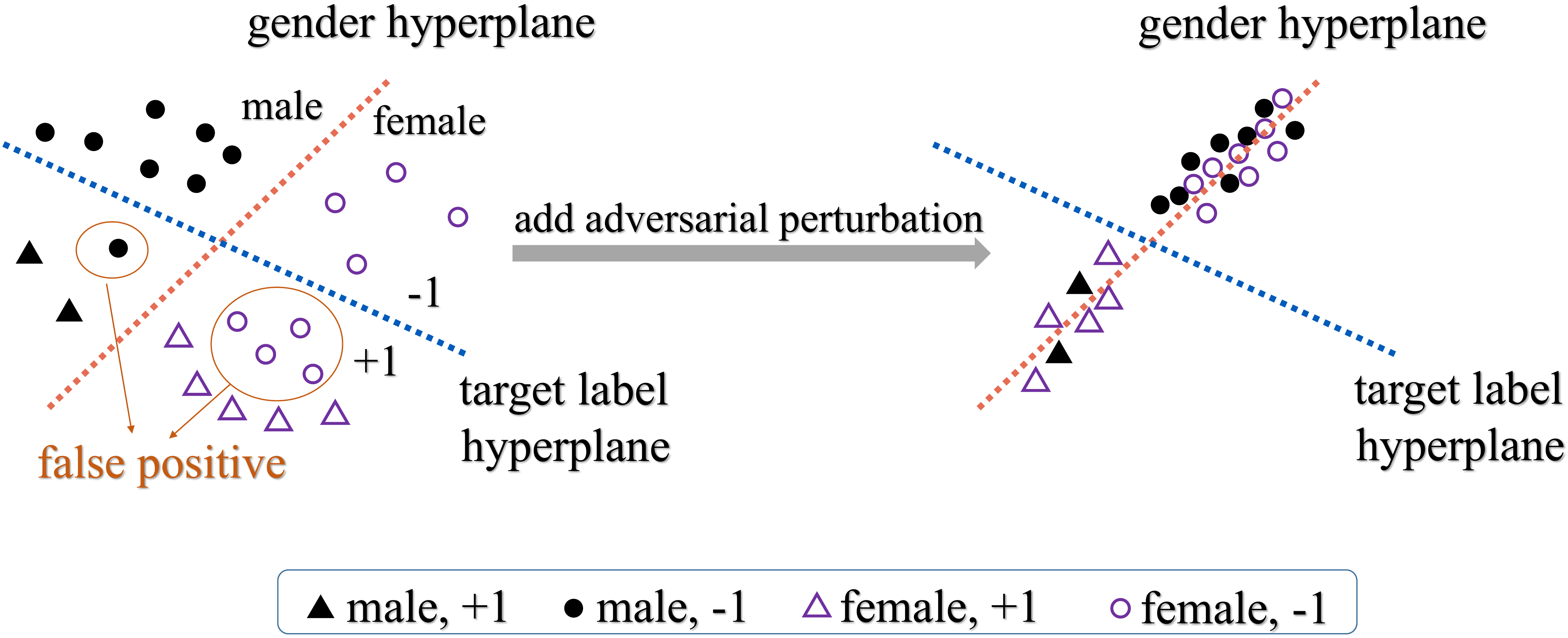}
\vspace{-2mm}
\caption{The basic idea of the proposed \algname{} (\algabbr). Gender bias exists in the left part, \ie, the false positive rate of females is much higher than males. Without adjusting the decision hyperplanes of the deployed model, \algabbr{} perturbs samples to decorrelate the target label and gender in latent space. In the right part, perturbed samples tightly distribute along the gender hyperplane, meanwhile, preserving the distinguishability along the target label hyperplane.}
\label{figure:fig9}
\vspace{-2mm}
\end{figure}

\subsection{Overview of FAAP}
\label{subsection:Overview of FAAP}
The unfairness could be caused by the bias in training sets (\eg, skewed data distribution) and/or loose constraints in the training process. All of these lead to spurious correlations between target labels and protected attributes, \eg, gender and ethnicity. In a dataset, females may have much more positive samples than males. As illustrated in Fig.~\ref{figure:fig9} (left), the model learns such spurious gender correlation so that the false positive rate of the target label varies significantly for males and females. Therefore, the key of mitigating unfairness is to break spurious correlations between target labels and protected attributes. 

In this paper, we propose the fairness-aware adversarial perturbation (FAAP) to mitigate model unfairness by hiding the information of protected attributes from the feature extraction process, so that the model would not correlate predictions with protected attributes. The basic idea is to leverage adversarial perturbation to remap the original samples to the position close to the decision hyperplane of the protected attribute in the latent space (\eg, on the surface of gender hyperplane in the figure). Note that the distinguishability of these perturbed samples along the original target label decision hyperplanes should be preserved, as shown in Fig.~\ref{figure:fig9} (right). In this way, the deployed model can not distinguish the protected attributes from the perturbed images during feature extraction. Therefore, the protected attribute would become uncorrelated to the target label.
In other words, the model would fairly treat samples with different protected attributes.

The pipeline of \algabbr{} is overviewed in Fig.~\ref{figure:fig3}, where there are two learnable components: 1) the generator that perturbs samples to regulate their distribution in the latent space, and 2) the discriminator that distinguishes the protected attribute. The deployed model is assumed to be a classification model that could be split into a feature extractor (\ie, from image to latent space) and a label predictor (\ie, from latent space to final label). Please note that we freeze the parameters of the deployed model. Sharing the spirit of general GANs during the training process, the discriminator is trained to distinguish the protected attribute from representations of the model, while the generator learns to fail the discriminator, thus synthesizing fairness-aware perturbation that reduces the information of the protected attribute in the latent representations.

\subsection{Loss Functions}
\begin{figure}[!t]
\centering
\includegraphics[width=1\linewidth]{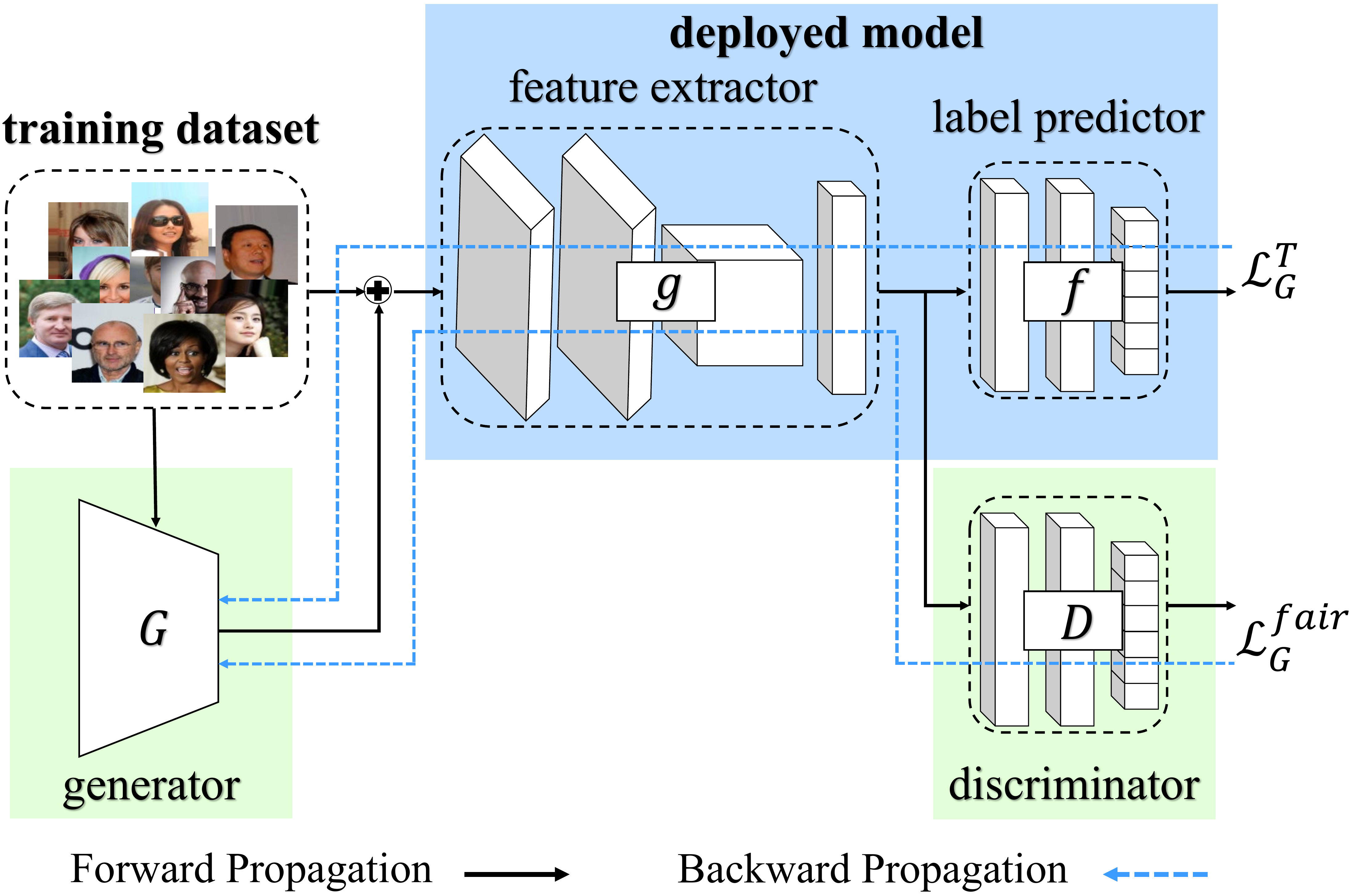}
\vspace{-2mm}
\caption{Overview of the proposed \algabbr, which consists of two learnable components, \ie, a generator for learning fairness-aware perturbation and a discriminator for distinguishing the protected attribute.}
\label{figure:fig3}
\vspace{-2mm}
\end{figure}

In this part, we detail the loss functions of the above-mentioned FAAP. As illustrated in Fig.~\ref{figure:fig3}, we assume a classification model that is divided into a feature extractor $g$ and a label predictor $f$. Given an input $x$, whose true label is $y$, the predicted label $\hat{y}=f(g(x))$. The generator $G$ generates perturbation based on input $x$ to obtain perturbed input $\hat{x}=x+G(x)$ subject to $\|\hat{x} - x \|_\infty \leq \epsilon$, and the discriminator $D$ is applied on the latent representations $\hat{r} = g(\hat{x})$ to distinguish a certain protected attribute $z$.

\textbf{Loss function of $D$:} Intuitively, with a deployed model, the unfairness is mainly caused by the feature extraction process which tends to correlate the protected attribute to those predicted in the target label, \ie, carrying distinguishable information from the protected attribute to the latent representations. Thus, the label predictor would utilize that distinguishable sensitive information to bias its final prediction. Based on the above hypothesis, we first need to let the discriminator $D$ aware of the protected attribute $z$ in the latent representation, \ie, perfectly predicting $z$. With such awareness, the generator $G$ is able to adversarially perturb inputs towards hiding the protected attribute in the latent representation. Therefore, the discriminator loss can be expressed as
\begin{equation}
\mathcal{L}_{D} = \mathcal{J}\left( D\left( g(\hat{x}) \right), z \right),
\label{eq:LD}
\end{equation}
where $\mathcal{J}(\cdot, \cdot)$ denotes cross-entropy, $\hat{x}$ is the perturbed input, and $z$ indicates the true label of the protected attribute. 

\textbf{Loss functions of $G$:} By contrast, the generator $G$ aims to fail $D$, and an intuitive solution is to maximize $\mathcal{L}_D$ on perturbed samples $\hat{x}$. However, this will push the latent representations towards the opposite side of the protected attribute, \eg, female flips to male. Therefore, we further let $D$ make random guess on the representation of $\hat{x}$, increasing entropy of the protected attribute on perturbed samples. The fairness loss can be written as
\begin{equation}
    \mathcal{L}_G^{fair} = - \mathcal{L}_D - \alpha\mathcal{H}\left(D\left( g(\hat{x}) \right)\right) ,
\label{eq:fair}
\end{equation}
where $\mathcal{H}(\cdot)$ calculates the entropy, $\alpha>0$ is a relatively small value controls the regularization of entropy loss. Besides $\mathcal{L}_G^{fair}$ that encourages fairness-aware perturbation, at the same time we need to preserve the model performance on the target label. The target label prediction loss is
\begin{equation}
\mathcal{L}_G^{T} = \mathcal{J}(f(g(\hat{x})), y).
\end{equation}

Above all, the total loss for generator $G$ in \algabbr{} consists of $\mathcal{L}_{G}^{fair}$ and $\mathcal{L}_G^{T}$, which can be summarized as the following
\begin{equation}
\mathcal{L}_G = \mathcal{L}_G^{fair} + \beta\mathcal{L}_G^{T}, 
\label{eq:allG}
\end{equation}  
where $\beta>0$ balances the performance of target label prediction and fairness.

\subsection{Training of \algabbr}
Based on Eq.~\ref{eq:LD} and Eq.~\ref{eq:allG}, in the training phase of \algabbr, the generator and the discriminator are optimized alternatively. The generator $G$ plays a mini-max game with $D$ where $D$ maximizes the ability to predict protected attribute $z$ while $G$ tries to minimize its ability. At the same time, $G$ tries to let $f$ still recognize the right target label for perturbed input data. Therefore,  the objectives of \algabbr{} can be formulated as follows:
\begin{equation}
\begin{aligned}
\arg\;\underset{G}{\max}\;\underset{D}{\min}\; & \mathcal{J}\left( D\left( \hat{r} \right), z \right) + \alpha\mathcal{H}\left(D\left( \hat{r} \right)\right) - \beta\mathcal{L}_G^{T}, \\
s.t.\; & \hat{r} = g(\hat{x}) = g\left( x + G(x) \right), \\
~ & \|\hat{x} - x \|_\infty \leq \epsilon,
\end{aligned}
\label{eq:train}
\end{equation}
where $D$ and $G$ are updated alternatively during the optimization. Please note that $\alpha$ is set to be 0 during updating $D$ to allow $D$ focus on distinguishing protected attributes. More detailed training algorithm of \algabbr{} can be found in Algorithm~\ref{algorithm:1}.

\begin{algorithm}[t]
	\caption{Training of \algabbr}
  	\label{algorithm:1}
  	\begin{algorithmic}
  		\Require Feature extractor $g$ and label prediction $f$ of a deployed model, loss weights $\alpha$ and $\beta$, learning rates $\eta_D$ and $\eta_G$,  maximum iteration $N$, and maximum perturbation magnitude $\epsilon$. The training images $x$, true labels $y$, and protected attribute labels $z$.
  		\Ensure Generator $G$
  		\State Initialize the generator $G$ and discriminator $D$.
  		\For{$i=1,\cdots,N$}
  			\State Get a batch of $n$ inputs $x_i$ and labels $y_i$ and $z_i$
  			\State Get perturbed inputs $\hat{x}_i=x_i + G(x_i)$
  			\State Clip $\hat{x_i}$ to meet $\|\hat{x_i} - x_i \|_\infty \leq \epsilon$
  			\State Get model feature $\hat{r_i} = g(\hat{x_i})$
			\State Calculate discriminator loss $$\mathcal{L}_D = \frac{1}{n} \sum_{i=1}^{n} \mathcal{J}\left( D\left( \hat{r_i} \right), z_i \right)$$
			\State Update $D \gets D - \eta_D \nabla_D\mathcal{L}_D$
			\State Calculate fairness loss $$\mathcal{L}_G^{fair} =-\frac{1}{n} \sum_{i=1}^{n}\left[\mathcal{J}\left( D\left( \hat{r_i} \right), z_i \right) + \alpha\mathcal{H}\left(D\left(\hat{r_i} \right)\right) \right]$$
			\State  Calculate target label prediction loss $$\mathcal{L}_G^{T} = \frac{1}{n} \sum_{i=1}^{n} \mathcal{J}\left(f(\hat{r_i}),y_i\right)$$
			\State Get total loss of $G$,  $\mathcal{L}_G =  \mathcal{L}_G^{fair} + \beta\mathcal{L}_G^{T} $
			\State Update $G \gets G - \eta_G \nabla_G\mathcal{L}_G$
		\EndFor
	\end{algorithmic}
\end{algorithm}

\section{Experimental Evaluation} \label{section:Experiments}
In this section, we first describe our experimental setup (Section \ref{subsection:Datasets and Experiments Setup}) . Then, we quantitatively (Section \ref{subsection:Diagnosis of Unfairness Mitigation}) and qualitatively (Section \ref{subsection:Visualization}) evaluate the proposed \algabbr{} on different deployed models. Finally, we investigate the transferability of adversarial perturbation generated by \algabbr{} on real-world commercial systems (Section \ref{subsection:Commercial APIs study}).

\subsection{Experimental Setup} \label{subsection:Datasets and Experiments Setup}

\noindent\textbf{Datasets.} We adopt two face datasets in our evaluation, \ie, CelebA\footnote{\url{http://mmlab.ie.cuhk.edu.hk/projects/CelebA.html}} and LFW\footnote{\url{http://vis-www.cs.umass.edu/lfw/}, attribute annotations are provided in~\cite{b36}},  which carry those commonly protected attributes like gender. The CelebA dataset consists of 202,599 images along with 40 attributes per image, and LFW has 13,244 images along with 73 attributes per image. We take gender as the protected attribute to measure the fairness of model prediction for target labels. In CelebA,  the \emph{Smiling}, \emph{Attractive}, and \emph{Blond\_Hair} are chosen as target labels. Similarly, \emph{Smiling}, \emph{Wavy\_Hair}, and \emph{Young} are selected as the target labels in LFW. We randomly divide the original training set of CelebA into two equal parts for training the deployed model and our \algabbr, respectively. For LFW, it is randomly split to get a 6k training set, a 3.6k validation set, and the rest as the testing set. For convenience, all the images are resized to 224$\times$224.

\noindent\textbf{Training details.}  To investigate the effectiveness of \algabbr{} in de-biasing models with different extent of unfairness, we train three kinds of models as the deployed models, \ie, normal training model, fair training model, and unfair training model. The normal training model is trained normally by minimizing the loss on target label. This kind of model will learn the intrinsic bias in the training dataset, \eg, the correlation between \emph{Smiling} and \emph{Male}. For the fair training model, we adopt adversarial training techniques~\cite{b56} to train a fair model, which maximizes the classifier’s ability to predict the target label, while minimizing the discriminator’s ability to predict the protected attribute. This kind of model has better fairness than the normal training model. To valid our method against more unfair models, which could be from malicious manipulations, \eg, data poison attack~\cite{b52} and malicious training~\cite{b53}, we apply two methods to amplify unfairness in deployed models. One is to flip labels (denoted as \textbf{LF}), \eg, randomly flipping the target labels. The other is to reverse the gradients of the discriminator in adversarial fair training (denoted as \textbf{RG}). These manipulations can strengthen the spurious correlation between target labels and gender.

For all deployed models, we use ResNet-18~\cite{b51} as the base architecture. We train all of these models for 30 epochs with a batch size of 64 using Adam optimizer with a learning rate of 5e-4. Once the training is finished, we fix the parameters of the deployed models. The generator $G$ in FAAP has a similar architecture with~\cite{b35}. Discriminator $D$ is connected to the last convolution layer of the feature extractor. To mitigate unfairness without harming the visual quality of a specific image, we set the maximum perturbation magnitude $\epsilon$ to 0.05.

\noindent\textbf{Evaluation metrics.} For fairness evaluation, we use the difference in demographic parity (DP) and difference in equalized odds (DEO) to evaluate model fairness. Meanwhile, the accuracy (ACC) of predicting target labels will also be reported. The DP calculates the absolute difference between the acceptance rates for each gender. A larger DP indicates that samples in the privileged group have higher chances to be predicted as positive than those in the unprivileged group. Ideally, the DP is equal to zero. By contrast, the DEO computes the absolute difference between the false negative rates and the false positive rates for each gender. A larger DEO means that samples in the privileged group have higher false positive rates and/or lower false negative rates than those in the unprivileged group. Therefore, the lower DEO the better.

\subsection{Quantitative Evaluation}
\label{subsection:Diagnosis of Unfairness Mitigation}

\begin{table*}[!ht]
    \begin{subtable}[b]{0.495\linewidth}
    \centering
\resizebox*{8cm}{!}{
\begin{tabular}{cccc}
\toprule
\textbf{Smiling}&{\textbf{ACC $\uparrow$}} &{\textbf{DP $\downarrow$}} &{\textbf{DEO $\downarrow$}} \\
\midrule
{Normal training} & {92.61\%} & {0.1748} & {0.0774} \\
\rowcolor{mygray}
{Normal training+FAAP} & {92.46\%} & {0.1426} & {0.0327} \\
\midrule
{Fair training} & {92.55\%} & {0.1275} & {0.0308} \\
\rowcolor{mygray}
{Fair training+FAAP} & {92.49\%} & {0.1326} & {0.0281} \\
\midrule
{Unfair training (LF)} & {91.48\%} & {0.2638} & {0.2737}\\
\rowcolor{mygray}
{Unfair training (LF)+FAAP} & {91.87\%} & {0.1268} & {0.0381}\\
\midrule
{Unfair training (RG)} & {91.76\%} & {0.2439} & {0.2306}\\
\rowcolor{mygray}
{Unfair training (RG)+FAAP} & {91.78\%} & {0.1321} & {0.0369}\\
\bottomrule
\end{tabular}
}
\caption{\label{table:big:sub1}Results on CelebA when the target label is \emph{Smiling}}
\end{subtable}
\vspace{0.1cm}
\begin{subtable}[b]{0.495\linewidth}
\centering
\resizebox*{8cm}{!}{
\begin{tabular}{cccc}
\toprule
\textbf{Smiling}&{\textbf{ACC $\uparrow$}}  &{\textbf{DP $\downarrow$}} &{\textbf{DEO $\downarrow$}}\\
\midrule
{Normal training} & {90.42\%}  & {0.3353} & {0.1472}\\
\rowcolor{mygray}
{Normal training+FAAP} & {89.80\%}  & {0.2910} & {0.0534}\\
\midrule
{Fair training} & {90.08\%}  & {0.2704} & {0.0318}\\
\rowcolor{mygray}
{Fair training+FAAP} & {88.75\%} & {0.2646} & {0.0136}\\
\midrule
{Unfair training (LF)} & {89.23\%}  & {0.3678} & {0.2340}\\
\rowcolor{mygray}
{Unfair training (LF)+FAAP} & {88.10\%}  & {0.3026} & {0.1076}\\
\midrule
{Unfair training (RG)} & {90.14\%}  & {0.3674} & {0.2257}\\
\rowcolor{mygray}
{Unfair training (RG)+FAAP} & {89.15\%}  & {0.2969} & {0.0782}\\
\bottomrule
\end{tabular}
}
\setcounter{subtable}{3}
\caption{\label{table:big:sub4}Results on LFW when the target label is \emph{Smiling}}
\end{subtable}
\vspace{0.1cm}
\begin{subtable}[b]{0.495\linewidth}
\centering
\resizebox*{8cm}{!}{
\begin{tabular}{cccc}
\toprule
\textbf{Attractive}&{\textbf{ACC $\uparrow$}} &{\textbf{DP $\downarrow$}} &{\textbf{DEO $\downarrow$}}\\
\midrule
{Normal training} & {82.43\%} & {0.5023} & {0.5683}\\
\rowcolor{mygray}
{Normal training+FAAP} & {79.73\%} & {0.2704} & {0.0621}\\
\midrule
{Fair training} & {79.56\%} & {0.2745} & {0.0724} \\
\rowcolor{mygray}
{Fair training+FAAP} & {79.31\%} & {0.2244} & {0.0434}\\
\midrule
{Unfair training (LF)} & {81.06\%} & {0.5566} & {0.7752}\\
\rowcolor{mygray}
{Unfair training (LF)+FAAP} & {79.08\%} & {0.2890} & {0.1179}\\
\midrule
{Unfair training (RG)} & {82.24\%} & {0.5547} & {0.7217}\\
\rowcolor{mygray}
{Unfair training (RG)+FAAP} & {79.37\%} & {0.2550} & {0.0539}\\
\bottomrule
\end{tabular}
}
\setcounter{subtable}{1}
\caption{\label{table:big:sub2}Results on CelebA when the target label is \emph{Attractive}}
\end{subtable}
\vspace{0.1cm}
    \begin{subtable}[b]{0.495\linewidth}
    \centering
\resizebox*{8cm}{!}{
\begin{tabular}{cccc}
\toprule
\textbf{Wavy\_Hair}&{\textbf{ACC $\uparrow$}}  &{\textbf{DP $\downarrow$}} &{\textbf{DEO $\downarrow$}}\\
\midrule
{Normal training} & {78.69\%} & {0.1707} & {0.1554}\\
\rowcolor{mygray}
{Normal training+FAAP} & {78.04\%} & {0.1241} & {0.0651}\\
\midrule
{Fair training} & {77.98\%} & {0.1337} & {0.0800}\\
\rowcolor{mygray}
{Fair training+FAAP} & {77.67\%} & {0.1094} & {0.0595}\\
\midrule
{Unfair training (LF)} & {78.35\%}  & {0.2383} & {0.2919}\\
\rowcolor{mygray}
{Unfair training (LF)+FAAP} & {77.19\%}  & {0.1765} & {0.1734}\\
\midrule
{Unfair training (RG)} & {77.59\%}  & {0.2724} & {0.3692}\\
\rowcolor{mygray}
{Unfair training (RG)+FAAP} & {77.10\%}  & {0.2128} & {0.2508}\\
\bottomrule
\end{tabular}
}
\setcounter{subtable}{4}
\caption{\label{table:big:sub5}Results on LFW when the target label is \emph{Wavy\_Hair}}
\end{subtable}
\vspace{0.1cm}
\begin{subtable}[b]{0.495\linewidth}
\centering
\resizebox*{8cm}{!}{
\begin{tabular}{cccc}
\toprule
\textbf{Blond\_Hair}&{\textbf{ACC $\uparrow$}} &{\textbf{DP $\downarrow$}} &{\textbf{DEO $\downarrow$}}\\
\midrule
{Normal training} & {95.63\%} & {0.1787} & {0.5299}\\
\rowcolor{mygray}
{Normal training+FAAP} & {94.52\%} & {0.1345} & {0.1013}\\
\midrule
{Fair training} & {94.41\%} & {0.1319} & {0.1587}\\
\rowcolor{mygray}
{Fair training+FAAP} & {94.05\%} & {0.1236} & {0.1043}\\
\midrule
{Unfair training (LF)} & {95.41\%} & {0.1733} & {0.6728}\\
\rowcolor{mygray}
{Unfair training (LF)+FAAP} & {94.49\%} & {0.1449} & {0.1321}\\
\midrule
{Unfair training (RG)} & {95.66\%} & {0.2041} & {0.6200}\\
\rowcolor{mygray}
{Unfair training (RG)+FAAP} & {94.26\%} & {0.1305} & {0.1209}\\
\bottomrule
\end{tabular}
}
\setcounter{subtable}{2}
\caption{\label{table:big:sub3}Results on CelebA when the target label is \emph{Blond\_Hair}}
\end{subtable}
\begin{subtable}[b]{0.495\linewidth}
\centering
\resizebox*{8cm}{!}{
\begin{tabular}{cccc}
\toprule
\textbf{Young}&{\textbf{ACC $\uparrow$}} &{\textbf{DP $\downarrow$}} &{\textbf{DEO $\downarrow$}}\\
\midrule
{Normal training} & {83.81\%} & {0.3511} & {0.5516}\\
\rowcolor{mygray}
{Normal training+FAAP} & {81.34\%} & {0.2281} & {0.2914}\\
\midrule
{Fair training} & {83.86\%} & {0.2500} & {0.2870}\\
\rowcolor{mygray}
{Fair training+FAAP} & {80.71\%} & {0.1515} & {0.1141}\\
\midrule
{Unfair training (LF)} & {83.04\%}  & {0.4813} & {0.8196}\\
\rowcolor{mygray}
{Unfair training (LF)+FAAP} & {80.40\%}  & {0.2550} & {0.3786}\\
\midrule
{Unfair training (RG)} & {83.72\%}  & {0.5002} & {0.8377}\\
\rowcolor{mygray}
{Unfair training (RG)+FAAP} & {82.30\%} & {0.1970} & {0.3048}\\
\bottomrule
\end{tabular}
}
\setcounter{subtable}{5}
\caption{\label{table:big:sub6}Results on LFW when the target label is \emph{Young}}
\end{subtable}

    \caption{Results of deployed models before and after embedded with the proposed FAAP on CelebA ( \Cref{table:big:sub1,table:big:sub2,table:big:sub3}) and LFW ( \Cref{table:big:sub4,table:big:sub5,table:big:sub6}). For fairness criterion \textbf{DP} and \textbf{DEO}, the lower the fairer. For accuracy \textbf{ACC}, the higher the better. }
    \label{table:big}
\vspace{-5mm}
\end{table*}

\Cref{table:big:sub1,table:big:sub2,table:big:sub3} show quantitative results of deployed models before and after embedded with the proposed FAAP on CelebA. We evaluate with three different target labels named \emph{Smiling}, \emph{Attractive} and \emph{Blond\_Hair} respectively with the protected attribute \emph{Male} (``+1'' in \emph{Male} means male and ``-1'' means female). Besides, we use three different kinds of models for each target label. As shown in Table~\ref{table:big}, there exists gender bias in normal training models, \eg, DP and DEO are larger than 0.5 when the target label is \emph{Attractive}. Fair training can get a fairer model by incorporating adversarial fairness techniques into training procedures. For instance, we can see in Table \ref{table:big:sub3}, fair training models have much lower DP (reduction from 0.5023 to 0.2745) and DEO (reduction from 0.5683 to 0.0724) than normal training models with a small drop in ACC (79.56\% comparing to 82.43\%). In contrast, unfair training amplifies gender bias and these models (LF and RG) show much more unfairness. For example, as shown in Table~\ref{table:big:sub1}, DP and DEO increase to about 0.25 with relatively high ACC (91.48\%, 91.76\% comparing to 92.61\% of normal training model).

We evaluate our method FAAP on the above deployed models. Not surprisingly, FAAP can improve fairness and maintain target label prediction accuracy for a deployed model. From \Cref{table:big:sub1,table:big:sub2,table:big:sub3}, we have the following observations. \textbf{(1) Normal training model.} For a normal training model, FAAP can improve its fairness and keep target label accuracy. We can see our method improves DP and DEO by 0.2319, 0.5062 respectively with accuracy loss less than 0.03 in Table~\ref{table:big:sub2}. 
\textbf{(2) Fair training model.} When adversarial fair training techniques are applied to the model training phase, our method can further improve the fairness of these models with slight accuracy drop, \eg, in Table~\ref{table:big:sub3}, FAAP still improve fairness (0.0083 and 0.0544 reduction in DP and DEO respectively) with slight accuracy degradation (from 94.41\% to 94.05\%).
\textbf{(3) Unfair training model.} For an unfair training model, FAAP can significantly improve its fairness with slight accuracy degradation. For instance, in Table~\ref{table:big:sub1}, FAAP can decrease DEO to about 0.04, maintaining ACC above 91\%.
\textbf{(4) Comparison between Normal training+FAAP and Fair training.} It is better to take model fairness into consideration in the training phase. However, in Table~\ref{table:big} we can see that a deployed normal training model embedded with FAAP can get comparable fairness performance as a fair training model (\eg, FAAP has even better DP and DEO in some cases) with almost the same accuracy (\ie, the difference in ACC is less than 0.3\% in most of the cases). For a deployed model, our method works after the training process without changing the model as compared to the fair training that needs to retrain or fine-tune the model. Similar observation can be observed in \Cref{table:big:sub4,table:big:sub5,table:big:sub6} on LFW dataset as well.

\subsection{Qualitative Evaluation}
\label{subsection:Visualization}
In this part, we further provide results of model explanation approaches Grad-CAM~\cite{b43} and T-SNE~\cite{b44} to better illustrate the effectiveness of our method.

\begin{figure}[!t]
     \centering
     \begin{subfigure}[b]{0.48\linewidth}
         \centering
         \includegraphics[width=1\linewidth]{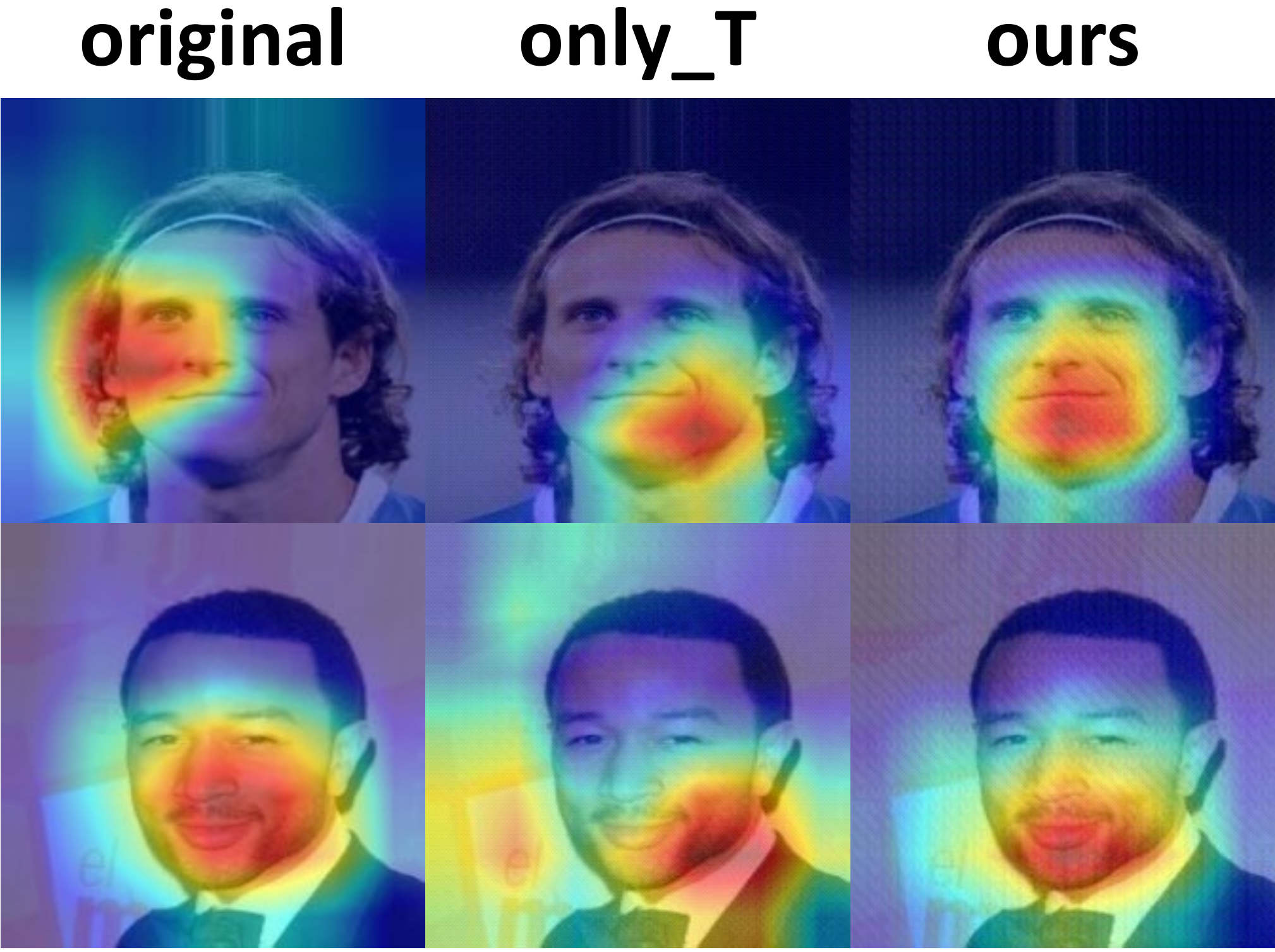}
         \caption{\label{figure:fig5:sub1}Normal training model}
         
     \end{subfigure}
     \hfill
     \begin{subfigure}[b]{0.48\linewidth}
         \centering
         \includegraphics[width=1\linewidth]{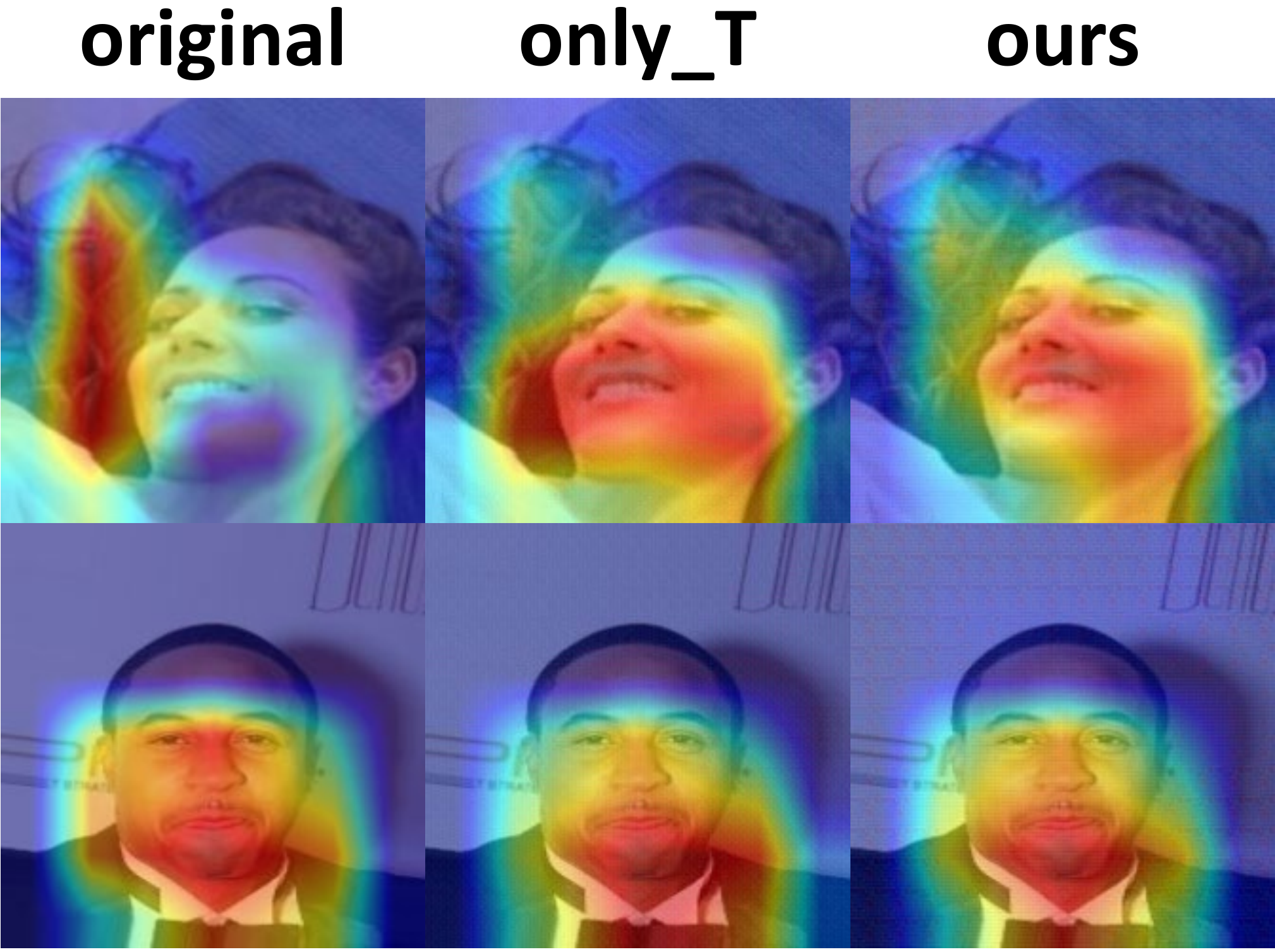}
         \caption{\label{figure:fig5:sub2}Fair training model}
         
     \end{subfigure}
     \hfill
     \begin{subfigure}[b]{0.48\linewidth}
         \centering
         \includegraphics[width=1\linewidth]{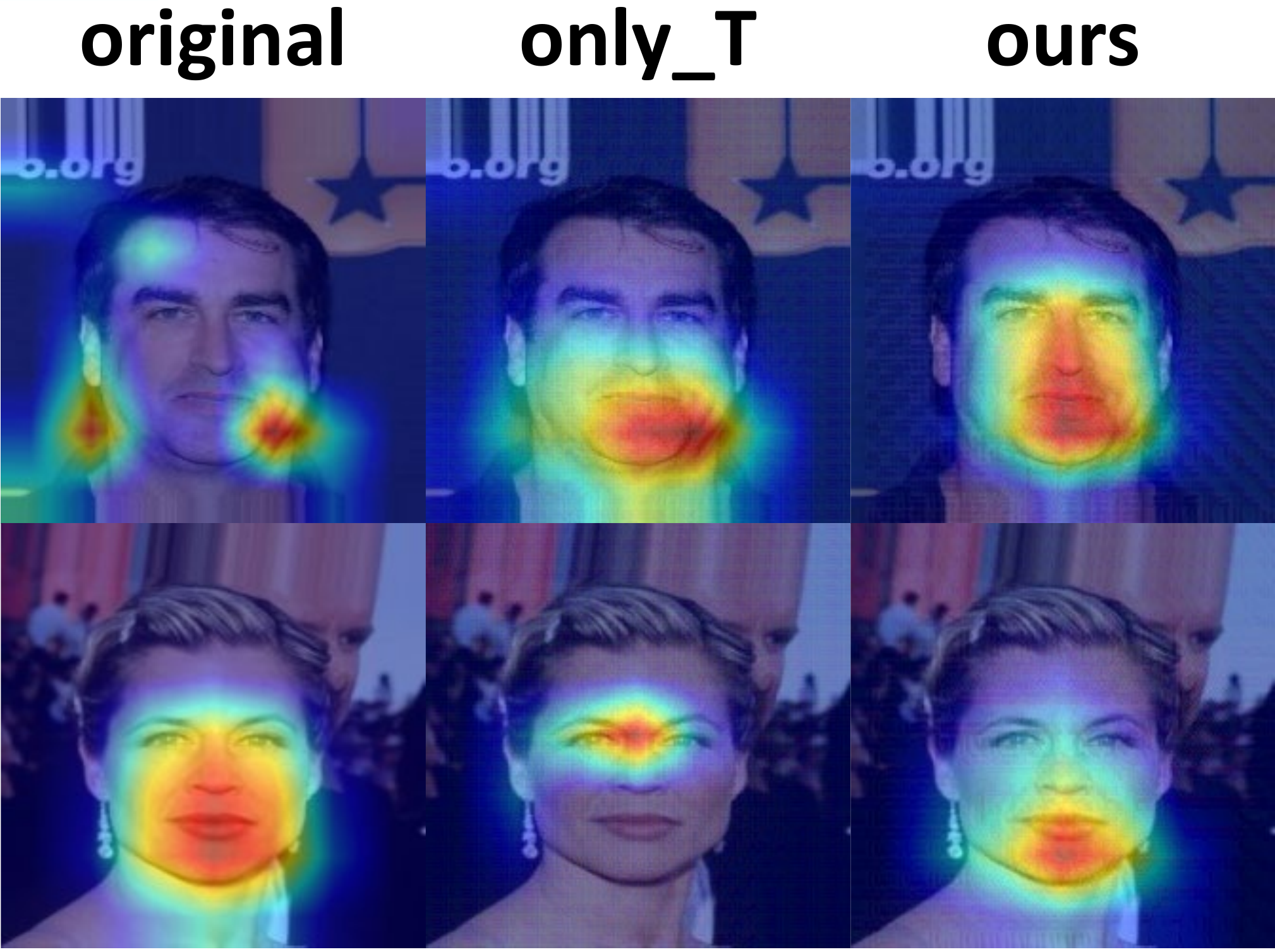}
         \caption{\label{figure:fig5:sub3}Unfair training model(LF)}
         
     \end{subfigure}
     \hfill
     \begin{subfigure}[b]{0.48\linewidth}
         \centering
         \includegraphics[width=1\linewidth]{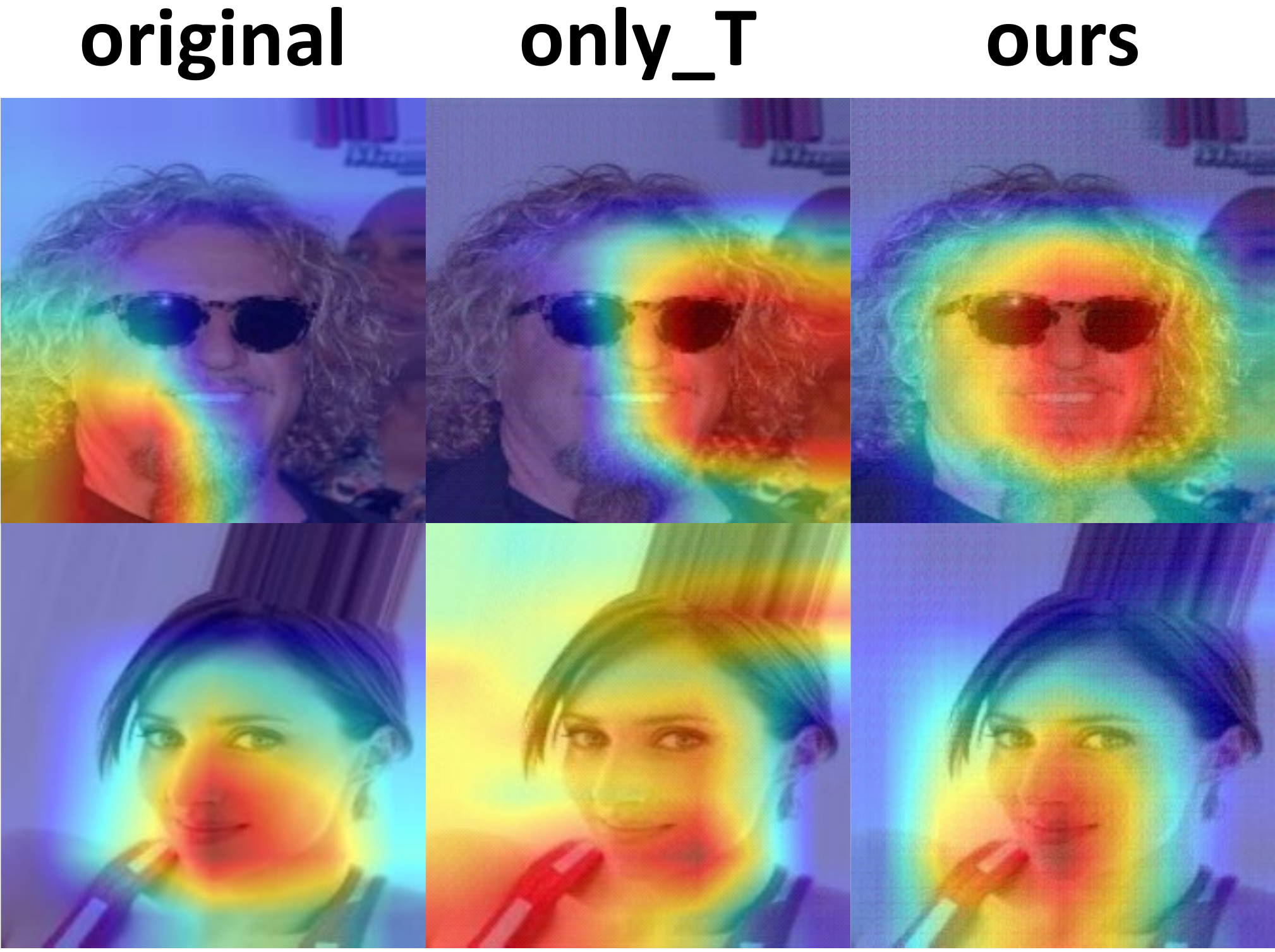}
         \caption{\label{figure:fig5:sub4}Unfair training model(RG)}
     \end{subfigure}
     
\caption{Grad-CAM results for three different models when the target label is \emph{Smiling} in CelebA. ``orginial'' denotes raw data, ``only\_T'' denotes images perturbed by $G$ which is only optimized on $\mathcal{L}_G^{T}$ without $\mathcal{L}_G^{fair}$, ``ours'' denotes images perturbed with fairness-aware adversarial perturbation generated by $G$ optimized on $\mathcal{L}_G$. (Better viewed in color)}
\label{fig:cam}
\vspace{-4mm}
\end{figure}

\textbf{Grad-CAM} is a model explanation method by visualizing the regions of input data that are important for predictions~\cite{b43}. We visualize a subset of test images that were originally false predicted by the deployed model but have been successfully recognized after perturbation in Fig.~\ref{fig:cam}. For each deployed model, we provide explanations on raw data, images perturbed by $G$ trained on $\mathcal{L}_G^{T}$ without $\mathcal{L}_G^{fair}$ and images perturbed by $G$ optimized on $\mathcal{L}_{G}$. 
\textbf{(1) Normal training model.} As shown in Fig.~\ref{figure:fig5:sub1}, for a normal training model, our adversarial perturbation can help the model focus on the right area (mouth) and make correct predictions. The red area of images in ``only\_T'' deviates little from the mouth. \textbf{(2) Fair training model.} Since this kind of model has less gender bias than other models, as shown in Fig.~\ref{figure:fig5:sub2}, $G$ optimized towards improving target label accuracy can get similar heat-maps as ``ours''. Both of them can help the deployed model focus on the right area. 
\textbf{(3) Unfair training model.} Unfair training models have larger gender tendency, thus we can see that perturbation in ``only\_T'' will let the model make correct predictions but mislead the model to focus on the unrelated area (\eg, eyes in Fig.~\ref{figure:fig5:sub3}, hair in Fig.~\ref{figure:fig5:sub4}). In contrast, our method helps the model focus on the right area and make right predictions.

\begin{figure}[!t]
     \centering
     \begin{subfigure}[b]{0.48\linewidth}
         \centering
         \includegraphics[width=1\linewidth]{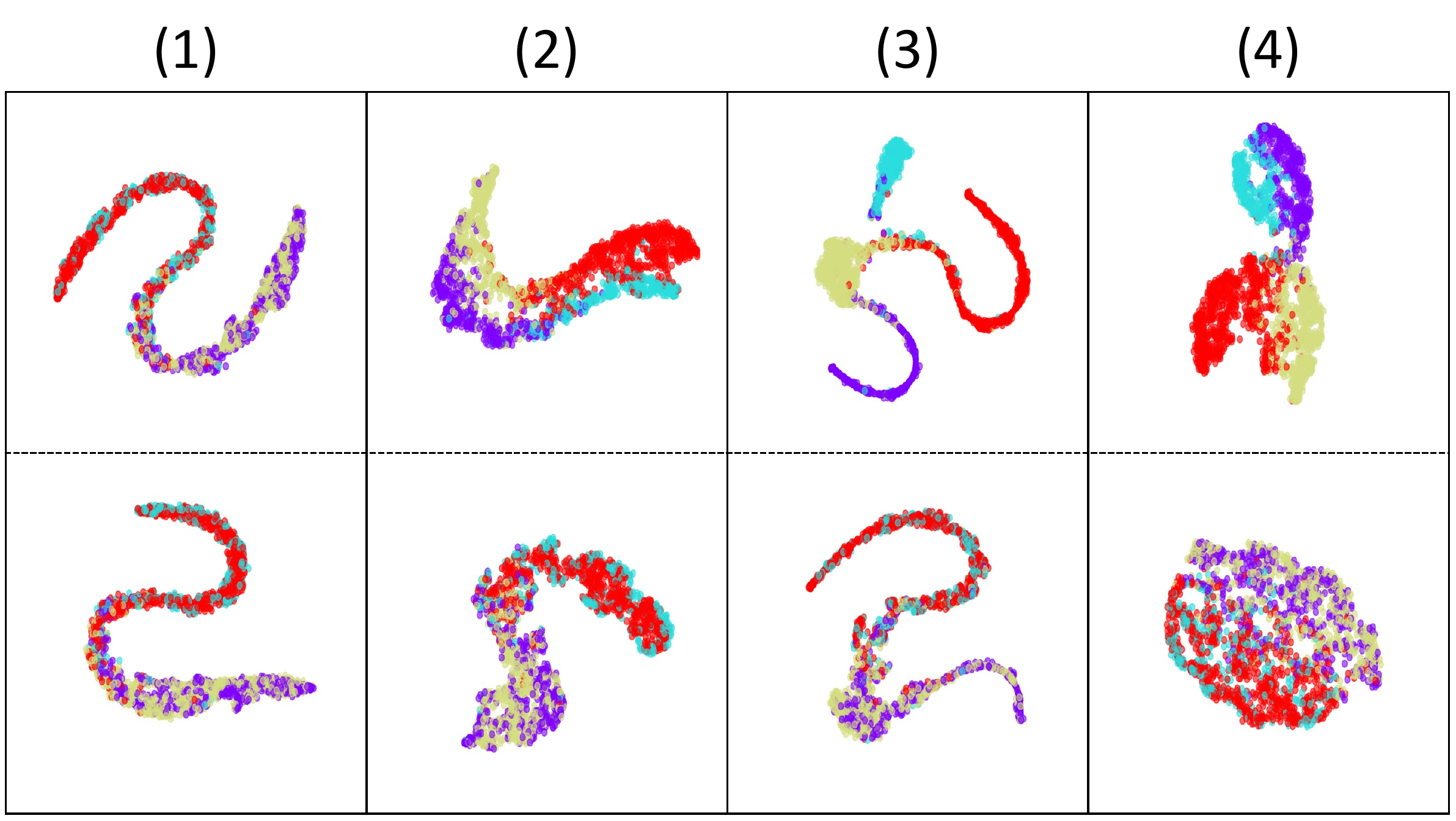}
         \caption{T-SNE for \emph{Smiling}}
         \label{figure:fig6:sub1}
     \end{subfigure}
     \hfill
     \begin{subfigure}[b]{0.48\linewidth}
         \centering
         \includegraphics[width=1\linewidth]{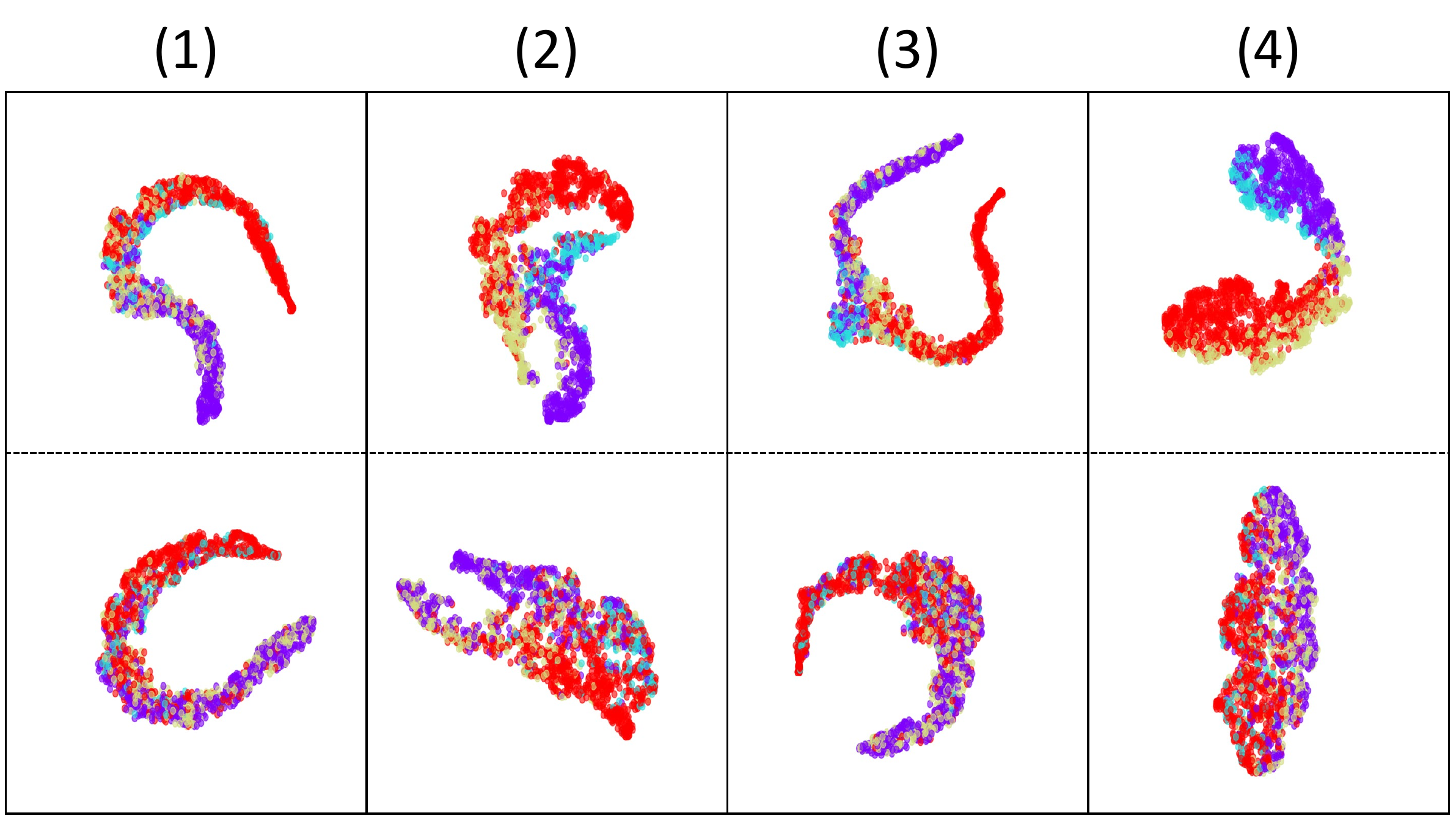}
         \caption{T-SNE for \emph{Attractive}}
         \label{figure:fig6:sub2}
     \end{subfigure}
     \hfill
     \begin{subfigure}[b]{0.48\linewidth}
         \centering
         \includegraphics[width=1\linewidth]{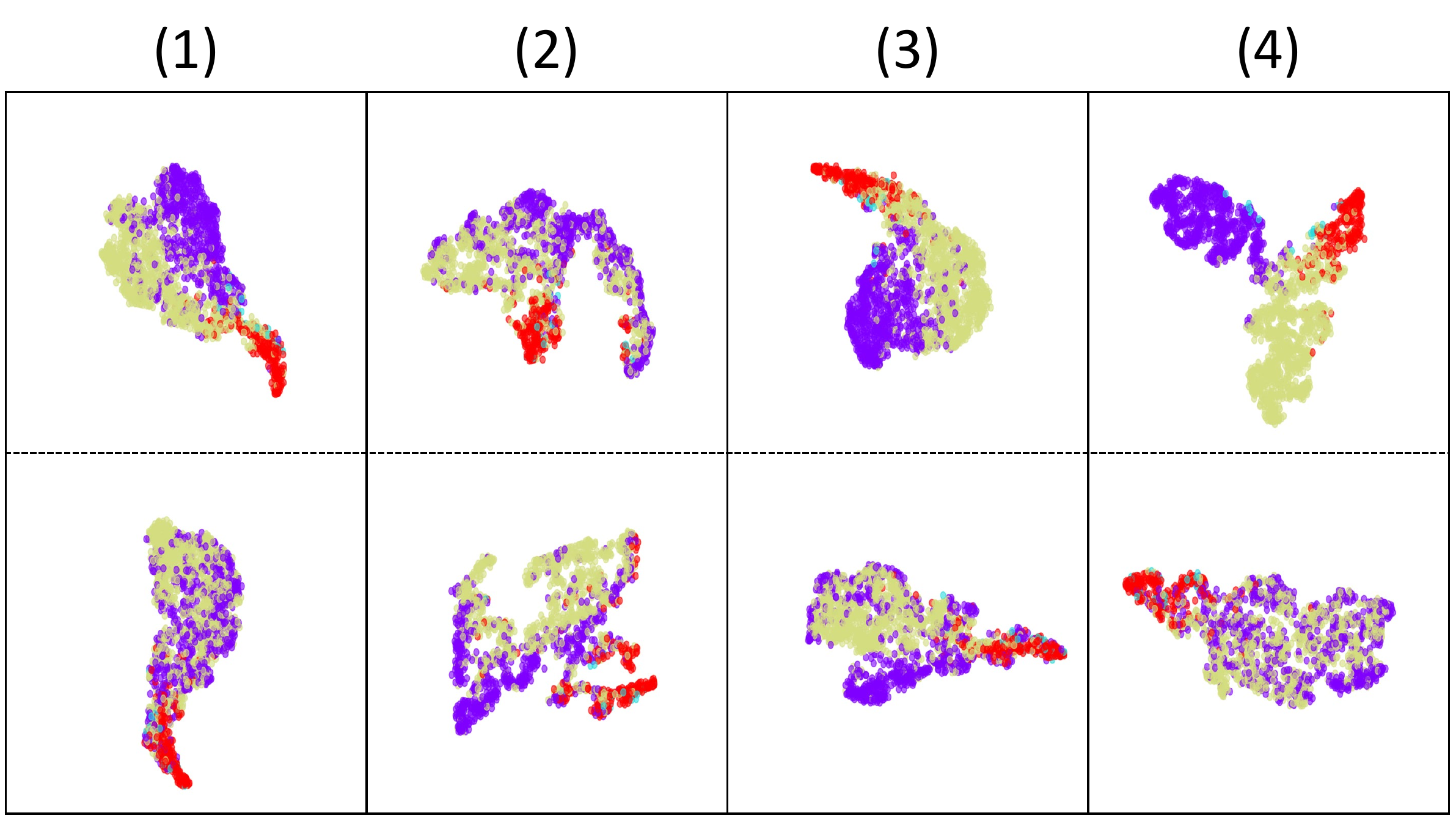}
         \caption{T-SNE for \emph{Blond\_hair}}
         \label{figure:fig6:sub3}
     \end{subfigure}
     \hfill
     \begin{subfigure}[b]{1\linewidth}
         \centering
         \vspace{2mm}
         \includegraphics[width=1\linewidth]{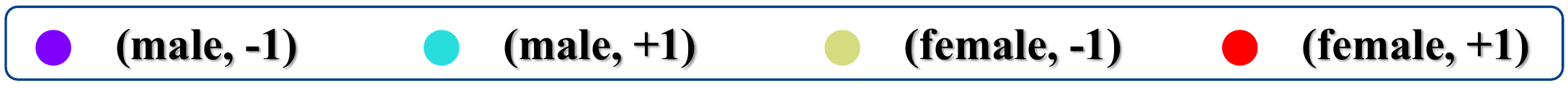}
         \label{figure:fig6:sub4}
         \vspace{-5mm}
     \end{subfigure}
\vspace{-4mm}
\caption{T-SNE results for three different models on \emph{Smiling}, \emph{Attractive} and \emph{Blond\_Hair} in CelebA. The upper row shows the results of the raw data and the bottom row shows the results of images perturbed with FAAP. In each sub-figure, the feature representation in column (1) is extracted from a normal training model, and column (2) from a fair training model, while column (3) from a LF model, column (4) from a RG model. (Better viewed in color)}
\label{fig:tsne}
\vspace{-2mm}
\end{figure}

\textbf{T-SNE} is a method to visualize high-dimensional data from a low dimension view. To better demonstrate that our method can hide sensitive information for images by remapping them close to the protected attribute decision hyperplane while maintaining the distance to the target label decision hyperplane in latent space of the deployed model, we utilize T-SNE to get low-dimensional embedding of data feature representation. More specifically, we extract feature vectors of these images with/without adversarial perturbation and visualize them in a two-dimension diagram with T-SNE. \textbf{(1) Normal training model.} From Fig.~\ref{figure:fig6:sub1} and Fig.~\ref{figure:fig6:sub2} we can see that for a normal training model, samples with different target labels for smiling and attractive classification are linearly separable in latent space, meanwhile, samples with different gender before and after perturbation are mixed. In Fig.~\ref{figure:fig6:sub3}, even the feature representations of samples (yellow and purple points) in normal training model are linearly separated by the protected attribute hyperplane when the target label is \emph{Blond\_Hair}, but FAAP can still effectively hide such sensitive information in latent feature space for samples. 
\textbf{(2) Fair training model.} Adversarial fair training can improve fairness, however, may slightly separate samples with different protected attributes (as shown in column (2) in Fig.~\ref{figure:fig6:sub1} and Fig.~\ref{figure:fig6:sub2}). In such a situation, our FAAP can make these samples become closer.
\textbf{(3) Unfair training model.} In unfair training models, feature representations of original images with different protected attributes are almost linear separable on the protected attribute hyperplane ( (male, -1) with (female, -1), (male, +1) with (female, +1) ). Once perturbed with adversarial perturbation generated by FAAP, samples with different gender become almost indistinguishable and mixed but they are well separable on the target label hyperplane. 

\subsection{Transferability of \algabbr}
\label{subsection:Commercial APIs study}
To demonstrate the transferability of adversarial perturbation generated by FAAP, we evaluate them on commercial face analyze APIs. At first, we investigate model fairness of these APIs in predicting ``smiling''. We upload testing dataset (about 20k images) from CelebA dataset to toady's commercial APIs, including Alibaba\footnote{\url{https://www.aliyun.com/}} and Baidu\footnote{\url{https://ai.baidu.com/}}. 
For Alibaba's face analyze API, it returns binary results in which ``0'' means ``not smiling'' and ``1'' means ``smiling''. For Baidu's face analyze API, it returns three categories named ``none'', ``smile'' and ``laugh''. We assume ``none'' means not smiling and others mean smiling. 
We find these APIs behave some extent of unfairness, \ie, DEO of them are about 0.1. Since this is a totally black-box scenario, we know nothing about the models behind these APIs. We try to train the generator with model ensemble techniques, taking the normal training model and the fair training model in Section~\ref{subsection:Diagnosis of Unfairness Mitigation} as surrogate models. Then we upload the perturbed images to these APIs and record results.

Table~\ref{table:tab7} shows the results of these face analyze APIs on original and perturbed images. From Table~\ref{table:tab7:sub1}, we can see that FAAP improves DP by $0.0293$ and decreases DEO to $0.0368$ with only $0.0026$ degradation in accuracy. Likewise, Table~\ref{table:tab7:sub2} shows $0.0411$, $0.0648$ improvement in DP and DEO while $0.0289$ degradation in accuracy for Baidu. 
These results show the transferability of FAAP and the potential usage of FAAP in black-box scenarios.

\begin{table}[htbp]
\begin{subtable}[htbp]{1.\linewidth}
\centering
\begin{tabular}{cccc}
\toprule
{\textbf{Alibaba's API}} & \textbf{ACC $\uparrow$} & \textbf{DP $\downarrow$} & \textbf{DEO $\downarrow$}\\
\midrule
{original images} & {90.20\%} & {0.1768} & {0.0952} \\
\rowcolor{mygray}
{after perturbation} & {89.94\%} & {0.1475} & {0.0368}\\
\bottomrule
\end{tabular}
\caption{\label{table:tab7:sub1}Results on Alibaba's face analyze API}
\vspace{2mm}
\end{subtable}

\begin{subtable}[htbp]{1.\linewidth}
\centering
\begin{tabular}{cccc}
\toprule
 {\textbf{Baidu's API}}& \textbf{ACC $\uparrow$} & \textbf{DP $\downarrow$} & \textbf{DEO $\downarrow$}  \\
\midrule
{original images} & {90.47\%} & {0.1817} & {0.1035} \\
\rowcolor{mygray}
{after perturbation} & {87.58\%} & {0.1406} & {0.0387} \\
\bottomrule
\end{tabular}
\caption{\label{table:tab7:sub2}Results on Baidu's face analyze API}
\vspace{2mm}
\end{subtable}

\vspace{-2mm}
\caption{\label{table:tab7}Performance on commercial face analyze APIs.}
\vspace{-4mm}
\end{table}

\section{Conclusion} \label{section:Conclusion}


This paper introduced the Fairness-Aware Adversarial Perturbation (\algabbr) to mitigate unfairness in deployed models. More specifically, \algabbr{} learns to perturb inputs, instead of changing the deployed models as the SOTA works, to disable deployed models from recognizing fairness-related features. To achieve this, we employed a discriminator to distinguish fairness-related attributes from latent representations of deployed models. Meanwhile, a generator was trained adversarially to deceive the discriminator, thus synthesizing fairness-aware perturbation that can hide the information of protected attributes. Extensive experiments demonstrated that \algabbr{} can effectively mitigate unfairness, \eg, improve DP and DEO by $27.5\%$ and $66.1\%$ respectively with only $1.5\%$ accuracy degradation on average for normal training models. 

In addition, evaluation on real-world commercial APIs showed significantly $19.5\%$ and $61.9\%$ improvement in DP and DEO with less than $1.7\%$ degradation in accuracy, which indicates the potential usage of the proposed \algabbr{} in the black-box scenario.
However, the black-box exploration is a side product of our current design. Since we assume to access the deployed models although do not modify them, it is still impractical for certain real cases like those commercial APIs. Therefore, we are considering future works on the black-box setting with more specific designs.

{\small
\bibliographystyle{ieee_fullname}
\bibliography{FAAP}
}

\end{document}